\documentclass[10pt]{article}

\usepackage[letterpaper,margin=0.68in]{geometry}
\usepackage{amsmath}
\usepackage{amssymb}
\usepackage{array}
\usepackage{booktabs}
\usepackage{graphicx}
\usepackage{microtype}
\usepackage{placeins}
\usepackage{balance}
\usepackage[numbers,sort&compress]{natbib}
\usepackage[hidelinks]{hyperref}

\newcommand{\workflowfigure}{%
  \begin{figure*}[t]
    \centering
    \includegraphics[width=0.96\textwidth]{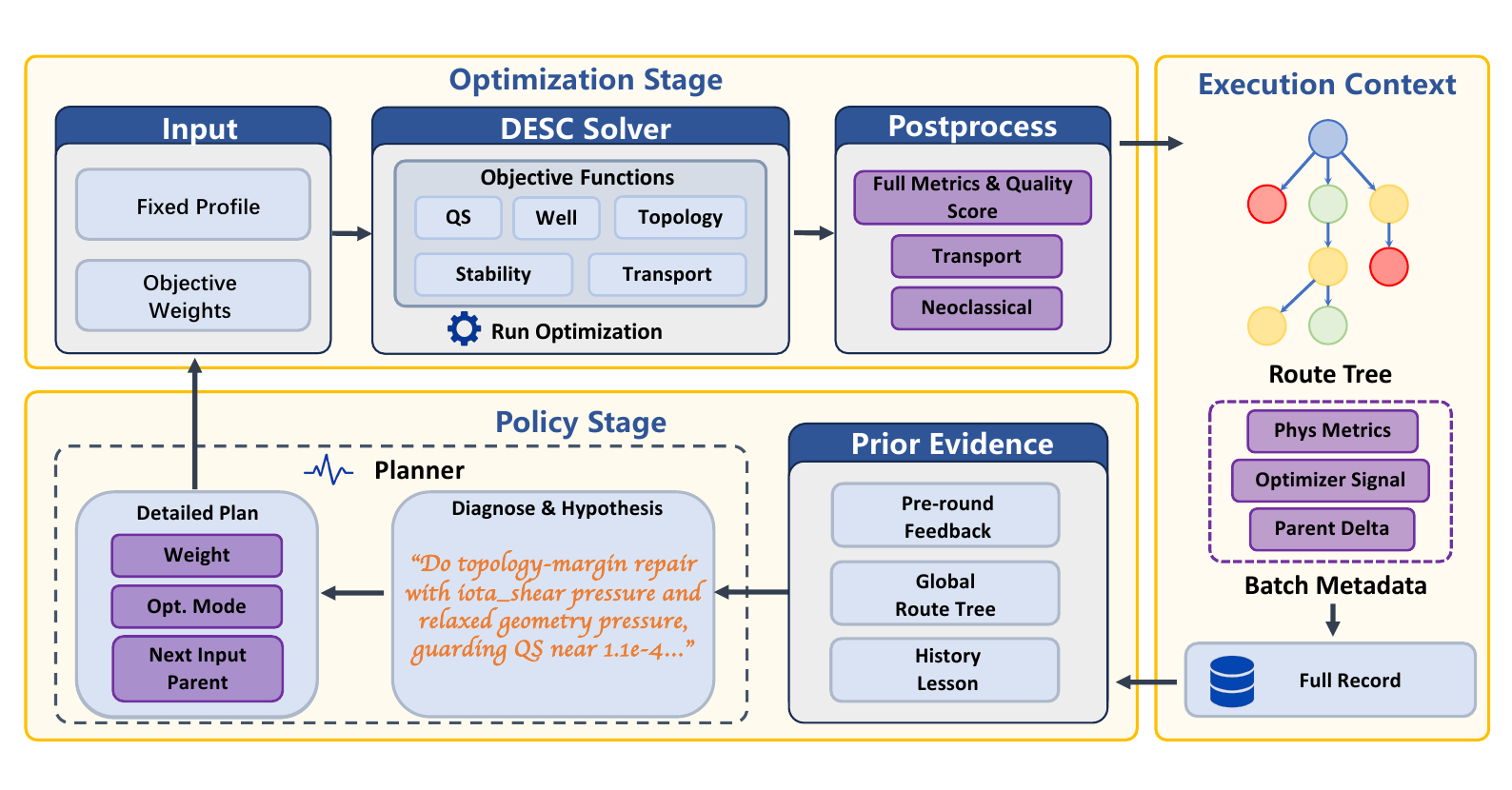}
    \caption{Agentic stage-one data-construction loop. A bounded agent selects
    an executable parent, Fourier-mode schedule, objective specification, and
    numerical budget. DESC performs the local optimization under a fixed physics
    contract, after which deterministic postprocessing records physical metrics,
    optimizer response, validity, and lineage. Every attempted action updates the
    configuration collection and the persistent transition corpus.}
    \label{fig:framework}
  \end{figure*}
}

\title{Agentic Stage-One Stellarator Optimization: Autonomous Multi-Objective Search for Finite-Beta Equilibria}
\author{Tingjia Zhang, Zhuoran Meng, Runlai Xu}
\date{August 1, 2026}

\begin{document}
\makeatletter
\twocolumn[
\begin{@twocolumnfalse}
\maketitle

\begin{abstract}
Stage-one stellarator design searches a high-dimensional family of
three-dimensional plasma boundaries and fixed-boundary MHD equilibria for
configurations that jointly meet requirements on confinement, field-line
topology, force balance, stability proxies, and geometry. These specifications do
not provide a general constructive map to a validated finite-beta equilibrium.
High-quality targets are commonly developed through iterative
numerical optimization whose outcome depends on the initial configuration,
active Fourier resolution, objective priorities, and local solver budget.
Coordinating this process is computationally costly and expert-intensive,
limiting both design throughput and the production of consistently evaluated
data. We present a proof of concept for \emph{agentic} stage-one optimization.
A bounded language-model agent diagnoses the current equilibrium and selects
the next local optimization experiment, while deterministic DESC execution
owns prescribed profiles and flux, symmetry, metric evaluation, solver
validity, and acceptance. On a common-budget subset from an expanding
finite-beta campaign, the number of gate-valid configurations increases from
five inputs to nineteen outputs; median Boozer QS RMS decreases from
$2.39\times10^{-4}$ to $1.07\times10^{-4}$, and median maximum principal
curvature decreases from $62.56$ to $33.00\,\mathrm{m}^{-1}$. A complementary
long route achieves a $9.10\times$ QS reduction while repairing magnetic-well
and curvature defects. The system also records every attempted local action as
transition evidence, yielding 734 structured parent--action--outcome records in
the reported experiments. These results show that agentic outer-loop control
can sustain finite-beta, multi-objective search and turn repeated optimization
into a scalable source of improved equilibria and reusable decision data.
\end{abstract}
\vspace{1em}
\end{@twocolumnfalse}
]
\makeatother

\section{Introduction}

Stage-one stellarator design is a constrained inverse problem over a chosen
family of three-dimensional plasma boundaries. In the conventional two-stage
workflow, each trial boundary defines a fixed-boundary ideal-MHD solve under a
specified field period, symmetry, profiles, and flux; an outer optimizer updates
the boundary, and a later stage designs coils that reproduce the target field
\cite{jorge2023single}. A useful target must jointly address confinement or
symmetry proxies, rotational-transform topology, force balance,
stability-related metrics, aspect ratio, and boundary regularity. The resulting
tradeoffs make stellarator design intrinsically multi-objective
\cite{bindel2023tradeoffs}.

The target metrics do not furnish a constructive inverse map to a validated
finite-beta equilibrium. Near-axis constructions and learned generators provide
useful candidate priors in restricted regimes
\cite{landreman2020well,cadena2025constellaration}; practical
finite-aspect-ratio targets are commonly refined through repeated equilibrium
solves and numerical optimization. The search is expensive and nonconvex, and
its outcome depends strongly on initialization, objective scalarization,
optimizer, and spectral resolution \cite{panici2026deflation}. Fourier
continuation improves robustness by activating boundary modes progressively,
while introducing further choices about how the search space should be expanded
\cite{jang2026spectral}.

These dependencies leave a sequential control problem outside the local
optimizer. Each solve updates boundary coefficients for a declared parent,
Fourier subspace, objective specification, and numerical budget. After observing
the physical and numerical response, an outer policy must decide whether to
continue, repair a degraded metric, expand the mode space, return to an earlier
parent, or explore another basin. The appropriate action changes along the
route, so a fixed scalarization or continuation schedule does not resolve this
decision problem. Managing the route therefore remains an expert-intensive
bottleneck on search throughput.

The same bottleneck shapes the available design data. Existing resources
broaden geometric coverage under specific contracts: QUASR assembles
vacuum-field coil configurations spanning QA and QH designs, whereas
ConStellaration targets QI equilibria and boundaries
\cite{giuliani2025quasr,cadena2025constellaration}. Using such configurations
for a finite-beta, multi-objective task requires equilibrium reconstruction,
reevaluation, and refinement. These computations reveal which interventions
improve, damage, or invalidate a candidate, information that endpoint
collections do not retain.

To address this coupled optimization-and-data problem, we introduce a bounded
agentic controller around deterministic DESC execution. At each epoch, the
controller selects a parent, Fourier-mode schedule, objective specification,
and local budget from the current route state. The harness enforces the
finite-beta physics contract, executes optimization, recomputes metrics, and
applies acceptance gates. Every attempted local solve is stored as
\emph{transition evidence}: its parent, declared action, optimizer response,
physical outcome, validity, and continuation decision. The resulting record
supports route-level credit assignment while accumulating reusable
state--action--outcome data. Figure~\ref{fig:framework} summarizes the
route-scale optimization and asynchronous multi-start system.

This paper makes three contributions:
\begin{itemize}
  \item \textbf{A sequential hyperparameter-control formulation.} We cast
  stage-one route control as a typed hyperparameter-optimization problem over
  parent selection, Fourier-mode access, objective specification, and local
  budget under an invariant finite-beta physics contract.

  \item \textbf{A physics-grounded agentic optimization system.} We combine a
  bounded language-model controller, deterministic DESC workers, metric-wise
  acceptance, persistent lineage, and cross-route memory to execute multi-start
  optimization routes automatically.

  \item \textbf{Optimization-driven data infrastructure.} We demonstrate
  repeated finite-beta, multi-objective improvement from heterogeneous source
  geometries and retain successful, rejected, and failed actions in a common
  transition schema. Each route contributes an improved design candidate and
  reusable evidence about the optimization process.
\end{itemize}

Together, these results establish a proof of concept at two connected levels:
an agent can execute physically meaningful stage-one routes, and the resulting
automation can scale those routes into a continuously expanding scientific data
resource. Comparative policy efficiency and learned transition models remain
separate questions.

\section{Background and problem formulation}
\label{sec:background}

\subsection{Fixed-boundary stage-one physics}

A stage-one calculation seeks a plasma boundary for which the magnetic field
is simultaneously close to a target symmetry, an accurate MHD equilibrium,
and compatible with geometric and stability requirements. For a prescribed
pressure profile $p(\rho)$, current profile, and toroidal flux, DESC represents
the ideal-MHD equilibrium through
\begin{equation}
  \nabla\!\times\!\mathbf{B}=\mu_0\mathbf{J},\qquad
  \mathbf{J}\!\times\!\mathbf{B}=\nabla p,\qquad
  \nabla\!\cdot\!\mathbf{B}=0,
  \label{eq:mhd}
\end{equation}
and solves these equations in a global spectral basis
\cite{panici2022desc1,dudt2022desc3}. The field-period number, stellarator
symmetry, helicity convention, transform branch, profiles, and flux define a
route-specific physics contract $\mathcal{C}$. They remain invariant while the
boundary is optimized.

For a stellarator-symmetric boundary, the design variables are Fourier
coefficients in
\begin{equation}
  \begin{aligned}
  R(\theta,\zeta)&=\sum_{m,n}R_{mn}
    \cos(m\theta-nN_{\mathrm{FP}}\zeta),\\
  Z(\theta,\zeta)&=\sum_{m,n}Z_{mn}
    \sin(m\theta-nN_{\mathrm{FP}}\zeta).
  \end{aligned}
  \label{eq:boundary}
\end{equation}
Low-order modes control global elongation, triangularity, axis displacement,
and other basin-scale geometry. Higher modes provide shorter-length-scale
corrections, while their derivatives can rapidly increase curvature and
coordinate distortion. Mode access therefore changes both the reachable
geometry and the conditioning of a local solve.

Deterministic postprocessing maps an equilibrium $x$ to
\begin{equation}
  \mathbf{m}(x)=\left(
  \epsilon_{\mathrm{QS}},W(\rho),\iota(\rho),r_F,A,
  \kappa_{\max},\mathbf{m}_{\mathrm{aux}}
  \right),
  \label{eq:metrics}
\end{equation}
where $\epsilon_{\mathrm{QS}}$ measures symmetry-breaking Boozer harmonics,
$W(\rho)$ is the dense magnetic-well profile, $\iota(\rho)$ is rotational
transform, $r_F$ is the normalized force residual, $A$ is aspect ratio, and
$\kappa_{\max}$ is maximum principal curvature. Quasisymmetry suppresses
symmetry-breaking guiding-center drifts; magnetic well provides a local
ideal-MHD stability proxy; transform controls field-line winding and resonance
topology; force residual checks consistency with the prescribed MHD state; and
aspect ratio and curvature guard geometric realizability. These quantities are
coupled through Eq.~\eqref{eq:mhd}, so improvement in one metric can consume
margin in another.
Boozer QS and magnetic-well diagnostics follow established definitions
\cite{landreman2022precise,landreman2020well,kim2020onaxis}.

The final design criterion is a conjunction of metric-wise gates
$g_j(\mathbf{m}(x))\leq 0$. This representation preserves the physical meaning
of each requirement. A scalar score is useful for ranking otherwise comparable
states, while acceptance remains determined by the complete gate vector.

\subsection{Sequential hyperparameter control}

DESC optimizes the physical boundary coefficients. We use
\emph{hyperparameter optimization} (HPO) for the agent's outer selection of
mode access, objective specification, and numerical budget, represented by
\begin{equation}
  h=(\mathcal{K},\mathbf{w},\boldsymbol{\eta},b),
  \label{eq:hpo}
\end{equation}
where $\mathcal{K}$ specifies the Fourier-mode schedule, $\mathbf{w}$ contains
objective weights, $\boldsymbol{\eta}$ contains typed objective parameters,
and $b$ is the numerical budget. Starting from parent $p$, the solve has the
schematic form
\begin{equation}
  \boldsymbol{\xi}^{\star}\approx
  \arg\min_{\boldsymbol{\xi}\in\Theta_{\mathcal{K}}(p)}
  \sum_r w_r\,\ell_r(\boldsymbol{\xi};\eta_r,\mathcal{C}),
  \label{eq:inner-objective}
\end{equation}
subject to equilibrium solution and numerical-validity checks. Here
$\Theta_{\mathcal{K}}(p)$ is the mode-restricted neighborhood of the parent.
The outer action is therefore
\begin{equation}
  a=(p,h)=(p,\mathcal{K},\mathbf{w},\boldsymbol{\eta},b).
  \label{eq:action}
\end{equation}
Parent and mode schedule are categorical decisions, cutoffs and budgets are
discrete, and weights and overrides are bounded numerical variables. The agent
configures this local experiment; the numerical optimizer determines the
Fourier update.

This HPO problem is state dependent. Applying the same $h$ to two parents can
reach different basins, trigger different solver behavior, or reverse a metric
tradeoff. The response surface is expensive, partially observed, and
discontinuous at failed solves and branch changes. Its objective also changes
with route state: a violated terminal gate calls for repair, whereas a
gate-valid state can prioritize lower QS while preserving well, topology,
geometry, and force margins. Temporary violations can expose a useful basin
when the damaged quantity remains repairable. Consequently, the outer policy
must reason over both the endpoint and the route by which it was produced.

\subsection{Transition evidence}

We represent one local experiment as
\begin{equation}
 e_i=\left(
  \mathrm{id}_p,\mathbf{m}(p),a_i,\mathrm{id}_{c},
  \mathbf{m}(x_i'),\Delta\mathbf{m}_i,
  \mathbf{s}_{i,\mathrm{opt}},v_i,d_i
 \right),
 \label{eq:transition}
\end{equation}
where $x_i'$ is the child when execution succeeds,
$\mathbf{s}_{i,\mathrm{opt}}$ records optimizer response, $v_i$ records solver
and evaluation validity, and $d_i$ indicates whether the child was selected for
continuation. Failed actions retain their parent, declared hyperparameters,
available optimizer trace, and failure class. The evaluated equilibria form
nodes in a route graph; transition records form directed edges.

Transition evidence exposes local controllability: which mode space and
objective emphasis produced a physical response from a particular parent. It
also separates an informative tradeoff from a uniformly poor endpoint. These
records are observations generated by a changing behavior policy, so they are
treated as confounded empirical priors rather than causal demonstrations. This
distinction informs both the agent context and the use of the accumulated data
for later learning.

\section{Agentic optimization method}
\label{sec:method}

\subsection{System overview}

The system separates HPO decisions from physical authority. The agent chooses
an existing parent and bounded values of Eq.~\eqref{eq:hpo}. The execution
harness validates the action, runs DESC, recomputes diagnostics, applies gates,
and writes provenance. Thus every policy decision becomes a declared and
replayable numerical experiment under the fixed contract $\mathcal{C}$.

\workflowfigure

As illustrated in Fig.~\ref{fig:framework}, an epoch has four stages:
\begin{enumerate}
  \item \textbf{Plan.} Construct a grounded route state and propose a small set
  of differentiated local hypotheses.
  \item \textbf{Execute.} Validate each action and evaluate independent DESC
  optimizations in parallel.
  \item \textbf{Evaluate.} Recompute all metrics on a common grid and retain
  endpoint, optimizer, validity, and failure observations.
  \item \textbf{Update.} Interpret the candidate set, select an executable next parent,
  and update compact route memory and persistent transition storage.
\end{enumerate}
The cycle continues until the route budget is exhausted. An outer asynchronous
scheduler runs multiple routes while candidate workers parallelize the physics
stage within each route.

\subsection{Grounded route state}

At epoch $t$, the planner receives a rendered state
\begin{equation}
 s_t=\{\mathcal{C},p_t,\mathbf{m}(p_t),\mathbf{g}(p_t),
 R_t,M_t,H_t,B_t\},
 \label{eq:agent-state}
\end{equation}
where $\mathbf{g}(p_t)$ contains signed gate margins, $R_t$ is a compact route
digest, $M_t$ is editable working memory, $H_t$ is retrieved cross-route
transition evidence, and $B_t$ is the explicit remaining budget. The route
digest is regenerated from stored artifacts and contains the active lineage,
recent epochs, and a curated parent menu. The menu includes the active parent,
recent successful children, earlier lineage anchors, and metric-specific
anchors for QS, well, transform, force, and aggregate quality. Every parent the
agent can name therefore resolves to an existing equilibrium artifact.

Working memory retains the current physical hypothesis, the dominant defect,
observed regressions, stagnation status, and the selected continuation parent.
Cross-route memory is a compact table of selected successful
parent--action--child
transitions rather than a transcript of prior reasoning. The prompt explicitly
labels these transitions as behavior-policy observations. This lets the agent
use repeated mechanisms as priors while requiring the current parent metrics
and gate margins to determine the action.

Budget is part of the state rather than an implicit conversation count. Opening
epochs permit acquisition and one diagnostic probe; development epochs compare
or repair mechanisms; exploitation concentrates on productive routes; closure
requires every candidate to be capable of producing the best feasible output.
Stagnation is tracked by parent, basin, and mechanism so repeated local polish
cannot consume the remaining route by default.

\subsection{Bounded planning and physics execution}

The planner follows a physics-first decision order. It identifies one primary
failed condition using its exact value and signed margin, records healthy
metrics as guards, assesses route phase and stagnation, and chooses an
acquisition, repair, escape, polish, validation, or backtracking mechanism. It
then emits schema-constrained JSON containing the parent identifier and, for
each candidate, its mode schedule, complete weight vector, sparse typed
overrides, iteration budget, force-solve budget, and rationale.

Candidates within an epoch test distinct hypotheses. At least one must make a
material intervention when the route is stagnant; early epochs admit at most
one small diagnostic probe, and closure epochs admit none. Exact weight bounds,
allowed override keys, solver caps, and parent identifiers are checked before
execution. Invalid plans are rejected before allocating a physics worker.

For valid candidate $i$, the deterministic layer evaluates
\begin{equation}
  (x_i',\mathbf{s}_{i,\mathrm{opt}},v_i)
  =\mathcal{D}(p_t,h_i;\mathcal{C}),
  \label{eq:execution}
\end{equation}
where $\mathcal{D}$ denotes equilibrium optimization followed by common-grid
postprocessing. Independent candidates run concurrently. Successful children
receive the complete metric vector and gate margins. The optimizer record
includes initial and final local cost, normalized cost change, iterations,
accepted steps, termination reason, and warnings; exceptions and partial traces
are retained for failed solves.

\subsection{Lesson and route update}

The update call has a narrower role than the planner. It receives the declared
plan, current-epoch outcomes, previous working memory, route budget, and the
curated executable parent menu. Long historical context is omitted at this
stage so that the observed epoch response remains the direct basis for credit
assignment. The agent summarizes whether the primary-defect hypothesis was
supported, identifies new debt, and selects the next parent from the menu.

Selection can continue a productive child, retain a material but gate-violating
tradeoff for repair, return to a lineage anchor, or terminate an exhausted
mechanism. The resulting memory stores the route-level lesson and high-level
next hypothesis. Concrete schedules, weights, and budgets are carried forward
only when the evidence makes them decisive; the next planning call otherwise
reconstructs an action from the new physical state. This separation prevents a
local lesson such as a strict mode prescription from becoming an unsupported
global policy.

\subsection{Persistent evidence and campaign scaling}

The implementation preserves execution at three resolutions. Per-route
directories contain equilibria, plans, logs, evaluations, failures, and complete
lineage. A relational table stores every attempted transition using
Eq.~\eqref{eq:transition}. Compact selected transitions are retrieved for agent
context. The full archive therefore remains available for analysis and learned
policies without expanding each online prompt.

A deterministic expert score
\begin{equation}
  Q(x)=\sum_j \alpha_j P_j\!\left(\mathbf{m}(x)\right)
  \label{eq:quality-score}
\end{equation}
provides a lower-is-better ranking anchor across QS, well, transform, geometry,
force balance, and solver health. Metric-wise gates retain authority over
acceptance. This division supports stable parent and endpoint ranking while
preserving every physical constraint explicitly.

At campaign scale, the scheduler advances independent routes asynchronously.
It samples a source lineage before selecting an eligible raw or improved member,
which limits repeated selection of a prolific family. Accepted outputs and the
best distance-to-goal result from incomplete routes can re-enter later
campaigns. Every completed route consequently contributes an optimized configuration,
the closest observed endpoint, and its complete transition evidence. This
mechanism turns repeated agentic optimization into a growing configuration and
state--action data resource under a shared evaluation contract.

\section{Experimental setup}
\label{sec:experimental-setup}

\subsection{Source configurations and execution}

The primary evaluation uses routes from an ongoing configuration-enhancement
campaign. Source boundaries are derived from QUASR coil-vacuum records
\cite{giuliani2025quasr} and prepared as finite-beta DESC equilibria with
pressure, Redl current, toroidal flux, and field normalization supplied before
agent execution. This preparation deliberately exposes each geometric prior to
a common class of finite-beta, multi-objective tests. Profiles and flux remain
fixed within each route.

We select 23 completed routes with a common eight-epoch budget for paired
evaluation. Every route starts from a distinct source record, and optimized
outputs are not recycled within this subset. A separate 50-epoch route provides
the temporal resolution needed to study acquisition, repair, and saturation.
An epoch contains one parent decision, four candidate local optimizations,
deterministic evaluation, and a next-parent update. The policy uses
GPT-5.6-sol with xhigh reasoning effort; calls receive the explicit state
described in Sec.~\ref{sec:method}. Independent routes and candidate workers
execute concurrently.

All reported configurations are stellarator-symmetric, positive-branch,
$N_{\rm FP}=2$ finite-beta QA equilibria with helicity $(1,0)$. Boundary
resolution is $M=N=6$ with radial resolution $L=10$; evaluation grids use
$L_{\rm grid}=20$ and $M_{\rm grid}=N_{\rm grid}=12$. The local objective
library contains Boozer QS, dense magnetic-well, rotational-transform,
aspect-ratio, curvature, and coordinate-quality terms. Magnetic-axis field
strength is retained as a low-priority scale diagnostic.

\subsection{Acceptance and paired evaluation}

Terminal acceptance requires Boozer QS RMS below $5\times10^{-4}$, dense-well
floor at least zero, well-positive fraction at least $0.9$, force RMS below
$10^{-4}$, maximum principal curvature below $33.2\,\mathrm{m}^{-1}$,
$0.25\leq|\iota_{\rm edge}|\leq1.2$, and $3\leq A\leq5$. Volume beta is reported
as an audit quantity because profiles and flux are fixed while geometry and
field strength evolve. Intermediate parents may carry a temporary well or
curvature violation; terminal outputs must satisfy every required gate.

Each campaign input is paired with one route output. Routes that reach
acceptance contribute their best accepted equilibrium under the deterministic
quality score in Eq.~\eqref{eq:quality-score}. The remaining routes contribute
the endpoint with minimum configured distance to the full gate. This preserves
a paired evaluation over the complete selected subset. The configuration gallery
shows the eight accepted pairs with the largest quality-score reductions.

For the detailed route, the reported output is the lowest-QS stored equilibrium
satisfying the well, force, curvature, iota, and geometry guards. Its deliberately
ambitious target is QS RMS below $10^{-6}$ with well-positive fraction at least
$0.98$ and $3\leq A\leq3.6$. The common-budget campaign subset contains 543
attempted transitions, and the long route contains 191, yielding 734 structured
observations across the two experiments.

\section{Experiments}
\label{sec:results}

The experiments test the two levels of the central claim. The multi-start
campaign asks whether the agentic workflow can repeatedly transform
heterogeneous geometric priors into better finite-beta, multi-objective
configurations under one evaluation contract. The retained transition corpus
then measures how much reusable optimization evidence this processing creates.
A long trajectory resolves the route-selection and repair behavior hidden by
the aggregate statistics.

\subsection{Multi-start finite-beta enhancement}

Only five inputs in the selected subset satisfy every terminal gate. After eight
epochs per source, 19 of the 23 selected outputs are gate-valid. Table
\ref{tab:multistart-summary} reports paired medians over all routes, including
the closest-to-gate endpoint from the four incomplete cases.

\begin{table}[!ht]
  \centering
  \caption{Paired medians for a common-budget subset of 23 distinct source
  configurations.}
  \label{tab:multistart-summary}
  \scriptsize
  \setlength{\tabcolsep}{4.0pt}
  \begin{tabular}{@{}lrr@{}}
    \toprule
    Metric & Input & Output \\
    \midrule
    Quality score $\downarrow$ & 18.95 & 9.48 \\
    QS RMS $\downarrow$ & $2.39{\times}10^{-4}$ & $1.07{\times}10^{-4}$ \\
    Well floor $\uparrow$ & 0 & 0 \\
    Well-positive fraction $\uparrow$ & 0.980 & 0.980 \\
    Max.\ curvature $\downarrow$ & 62.56 & 33.00 \\
    Force RMS $\downarrow$ & $4.91{\times}10^{-5}$ & $4.00{\times}10^{-5}$ \\
    \bottomrule
  \end{tabular}
\end{table}

The deterministic multi-objective score improves in 20 routes. QS RMS decreases
in 22 routes, with its median falling from $2.39\times10^{-4}$ to
$1.07\times10^{-4}$. Maximum curvature decreases in 20 routes and its median
moves from $62.56$ to $33.00\,\mathrm{m}^{-1}$, close to the terminal gate.
Force RMS improves in 16 routes. Ten routes improve both the dense-well floor
and positive fraction; their medians remain unchanged because many source
profiles already have a zero floor and $0.98$ positive fraction. Every selected
output preserves the prescribed iota and aspect-ratio intervals. The aggregate
shift therefore reflects joint finite-beta feasibility rather than QS reduction
alone.

\begin{figure*}[t]
  \centering
  \includegraphics[width=0.98\textwidth]{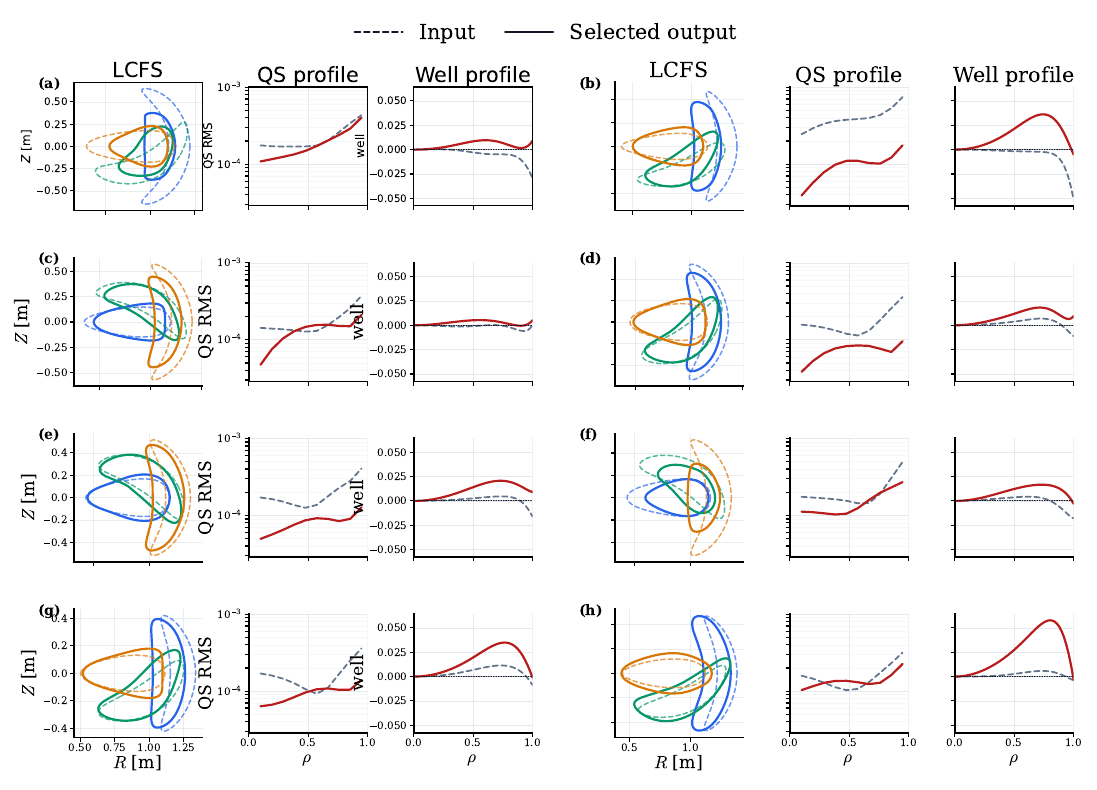}
  \caption{Input--output comparisons for the eight accepted configurations
  with the largest quality-score reductions. Each sample contains equally sized
  LCFS, radial Boozer-QS-RMS, and dense magnetic-well panels. Dashed curves
  denote inputs and solid curves denote selected outputs. In the LCFS panels,
  blue, green, and orange denote $\phi=0$, $\pi/(2N_{\rm FP})$, and
  $\pi/N_{\rm FP}$, respectively.}
  \label{fig:multistart-comparison}
\end{figure*}

Figure \ref{fig:multistart-comparison} shows that the improvements occur across
several LCFS families. The paired profiles also expose the central
multi-objective behavior: QS reductions coexist with preservation or repair of
the dense magnetic well, while substantial curvature violations are removed
through changes that remain moderate at the scale of the LCFS.

\subsection{Transition-data production}

The same short routes produce 543 attempted transitions. Each row couples a
parent metric state and executable action to a child metric state or failure,
parent-relative changes, optimizer feedback, validity, and continuation
decision. The records include children rejected by terminal gates, alternatives
that lost the within-epoch comparison, and solves that terminated before a
valid equilibrium. These outcomes locate tradeoff surfaces and numerical
failure boundaries that are absent from the selected endpoints.

The compact successful subset is retrieved by later agent calls as cross-route
experience; the complete corpus remains available for offline analysis. Its
fixed schema also admits learned state--action predictors, proposal priors, and
policy evaluation as physical and action-space coverage increases. The present
experiments use retrieval rather than a trained dynamics model. Crucially, each
additional agent route enlarges both products under the same provenance
contract: an evaluated configuration collection and an aligned transition
corpus. Agentic execution therefore converts optimization compute into a growing
record of designs, decisions, responses, and failures.

\subsection{Mechanism study along a long route}

The aggregate experiment establishes repeated configuration enhancement. A
50-epoch route reveals how one output is obtained when the source has competing
defects. Its input has $\beta_{\rm vol}=2.039\%$, QS RMS
$3.782\times10^{-4}$, dense-well minimum $-2.267\times10^{-2}$, maximum
principal curvature $50.997\,\mathrm{m}^{-1}$, and force RMS
$5.68\times10^{-5}$. The initial force solve is valid, while well and curvature
require substantial repair.

\begin{figure*}[t]
  \centering
  \includegraphics[width=0.97\textwidth]{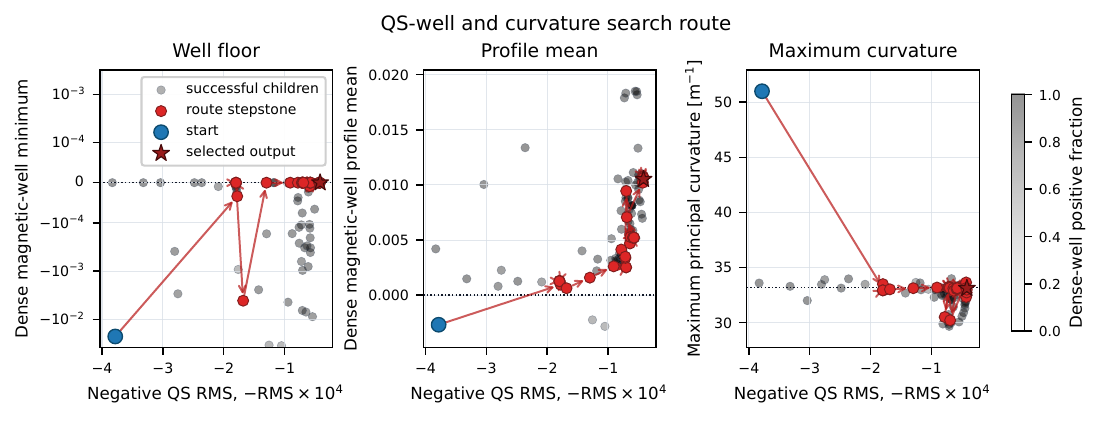}
  \caption{Executed search path in QS--well--curvature space. Gray points are
  successful children with QS RMS below $5\times10^{-4}$; red arrows show the
  selected parent--child ancestry, and the star marks the selected output. The
  left and middle panels show dense magnetic-well floor and profile mean. The
  right panel shows maximum principal curvature, with the dotted line marking
  the $33.2\,\mathrm{m}^{-1}$ gate. Gray intensity encodes dense-well positive
  fraction.}
  \label{fig:main-policy-route}
\end{figure*}

\begin{figure}[!t]
  \centering
  \includegraphics[width=0.98\linewidth]{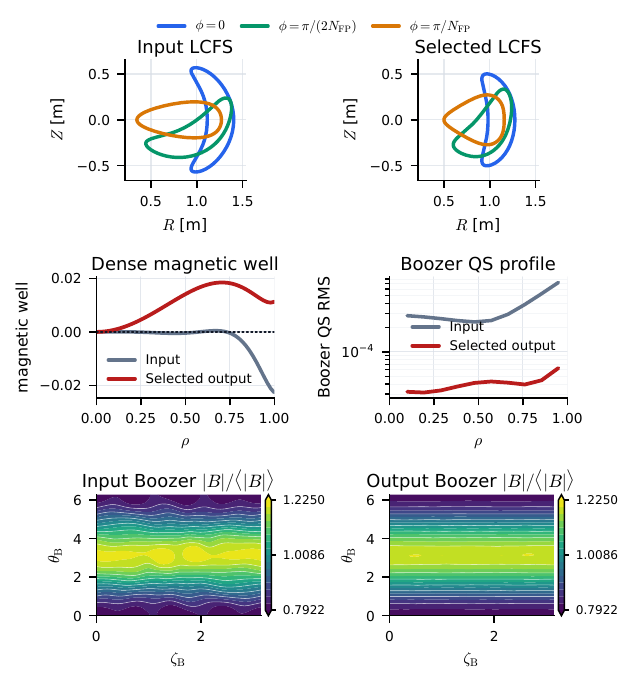}
  \caption{Input--output physics comparison for the long route. The top row
  compares LCFS cuts at three toroidal angles, the middle row compares dense
  magnetic-well and radial Boozer-QS-RMS profiles, and the bottom row shows
  normalized Boozer-coordinate field strength with common contour levels.}
  \label{fig:main-physics-comparison}
\end{figure}

\begin{figure}[!t]
  \centering
  \includegraphics[width=0.98\linewidth]{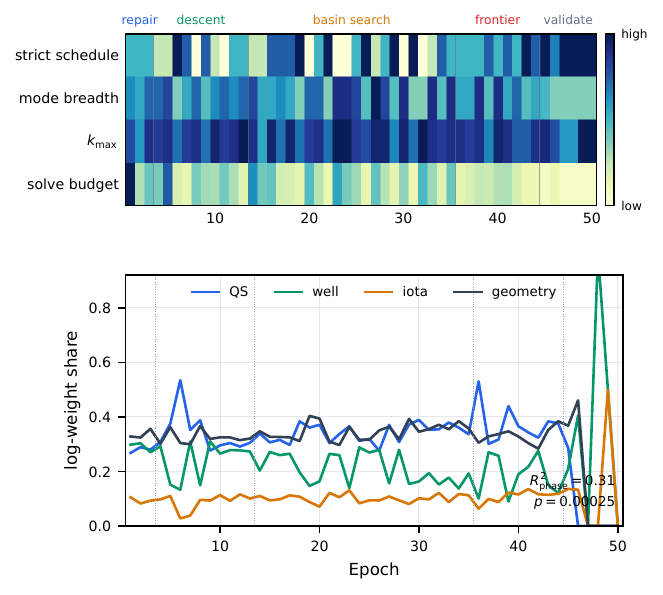}
  \caption{Executable action variation over 50 epochs. The upper panel
  summarizes mode schedule, mode breadth, maximum cutoff, and solve budget. The
  lower panel shows log-weight emphasis for QS, well, iota, and geometry
  objective groups. Phase labels are retrospective partitions defined by
  physical route events.}
  \label{fig:policy-phase-diagnostics}
\end{figure}

\begin{figure}[!t]
  \centering
  \includegraphics[width=0.98\linewidth]{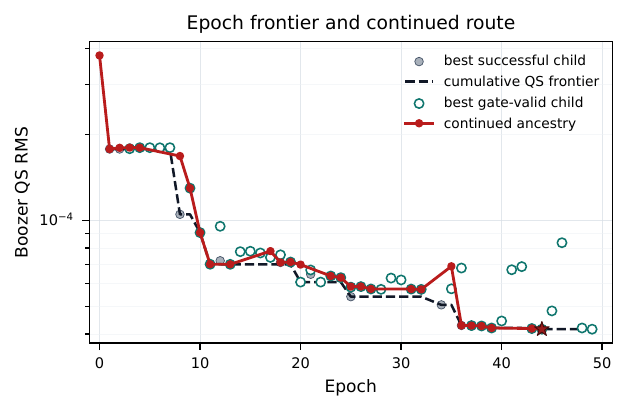}
  \caption{QS frontier and selected continuation route. Gray points show the
  best successful child in each epoch, the dashed line is the cumulative
  best-QS frontier, open circles identify the best gate-valid child, and the red
  line is the ancestry of the selected output.}
  \label{fig:epoch-frontiers-route}
\end{figure}

Figures \ref{fig:main-policy-route} and \ref{fig:main-physics-comparison} show
three stages. Epochs 1--3 first construct a valid working parent: QS falls to
$1.78\times10^{-4}$, the well floor rises to near zero, curvature falls below
the gate, and the positive well fraction reaches $0.98$. Guarded QS descent then
reaches $7.02\times10^{-5}$ by Epoch 11.

The subsequent improvement is nonmonotone. Epoch 17 accepts a temporary QS
increase while reducing curvature to $30.49\,\mathrm{m}^{-1}$ and increasing
the well mean. A shallow well defect introduced at Epoch 25 is repaired at
Epoch 26 with little QS cost. Epoch 35 makes a second displacement toward lower
curvature and stronger well behavior. From that parent, Epoch 36 acquires QS
$4.284\times10^{-5}$ with curvature slightly above the gate, and Epoch 37
repairs curvature while retaining the newly acquired QS level. These events
illustrate why continuation value cannot be inferred from instantaneous QS
rank.

The selected Epoch 44 equilibrium reaches QS RMS $4.157\times10^{-5}$, a
$9.10\times$ reduction from the input. Its dense-well floor is zero,
well-positive fraction is $0.980$, force RMS is $3.12\times10^{-5}$, maximum
curvature is $33.111\,\mathrm{m}^{-1}$, aspect ratio is $3.444$, and
$|\iota_{\rm edge}|=0.340$. Volume beta changes from $2.039\%$ to $1.832\%$
under fixed profiles and flux. The remaining six epochs vary parent replay,
mode support, and step scale without finding a lower-QS gate-valid state,
providing an empirical local-saturation test. The final QS remains above the
ambitious $10^{-6}$ target.

Figures \ref{fig:epoch-frontiers-route} and
\ref{fig:policy-phase-diagnostics} connect the physical path to the executed
actions. The policy descriptor combines log objective-group weights,
mode-schedule semantics, mode breadth and cutoff, solver budgets, and explicit
well guards. After standardization, the retrospective physical phases explain
$R^2_{\rm phase}=0.31$ of descriptor variance; a label-permutation test gives
$p=2.5\times10^{-4}$. This descriptive result confirms that the executed action
distribution changes across repair, descent, acquisition, and saturation
segments. The route contributes 191 transition records that preserve these
changes at candidate resolution.

\section{Discussion}

\subsection{From agentic optimization to data infrastructure}

The significance of the agentic layer extends beyond replacing manual route
decisions. It makes repeated, stateful optimization across heterogeneous source
configurations an executable scientific process. Source collections can provide
broad geometric coverage even when their original field model and objective set
differ from the target task. The workflow supplies a consistent finite-beta
equilibrium, metric, and acceptance layer, then lets an agent improve each
source under that layer. In the selected evaluation subset, this process
increases gate-valid coverage and shifts QS, curvature, force balance, and
composite quality together.

This capability suggests a practical route toward foundational data
infrastructure for stellarator design. Continued execution can extend the
configuration collection across source families, field periods, helicities,
profile regimes, and acceptance contracts. Because the physics and provenance
schema remains stable, the resulting configurations become comparable even as
the campaign and agent policy evolve.

Transition evidence further compounds the scientific value of the same solver
budget. Endpoint databases support design retrieval and generative modeling;
the transition corpus samples local responses to mode access, objective
emphasis, and numerical scale. These records can support action-conditioned
surrogates, learned proposal priors, failure prediction, and offline policy
comparison once coverage is sufficient. Lineage and rejected alternatives are
especially important because they retain the local tradeoffs that selected
designs alone cannot reconstruct. Agentic optimization thereby offers a path
from isolated expert runs to cumulative, machine-usable optimization knowledge.

\subsection{Interpretation of the long route}

The detailed trajectory supplies the route-level proof of concept for agentic
control. The selected ancestry is non-greedy in QS, temporarily spends well or
curvature margin, repairs the resulting debt, and changes executable actions as
the frontier evolves. Candidate-level records make these events observable.
They also distinguish frontier acquisition from late local saturation, where
additional mode schedules and parent replays produce no better valid state.

The action analysis is observational. Parent choice, objective weights, mode
support, and budget often change jointly, so the present route does not identify
a causal transition law or establish language-model sample efficiency. Its role
is to show that an agentic controller can sustain a physically valid,
multi-stage route and that the retained representation resolves the relevant
state--action changes.

\subsection{Scope and next steps}

The reported evaluation covers finite-beta, $N_{\rm FP}=2$ QA equilibria under
one policy and one principal source family. Matched-budget static, greedy,
random, and memory-ablated controllers are required to quantify policy
efficiency. Broader campaigns should balance geometry clusters because distinct
source records can occupy nearby boundary families. Four short routes remain
outside the terminal gate, and the long route does not reach its $10^{-6}$ QS
target. Magnetic-axis field strength is currently a soft scale objective, so
reactor-normalized comparisons require stricter scaling control. Coil
realizability and engineering loads
\cite{kaptanoglu2024dipole,kaptanoglu2025passive}, together with
energetic-particle confinement, turbulence, and transport, remain downstream
tests for the fixed-boundary outputs.

These limitations define a direct expansion path. The existing scheduler can
increase physical and geometric coverage, while the persistent schema allows
new policy versions and learned priors to train on the same accumulated
transitions. At sufficient scale, this creates a progression from retrieved
examples to learned state--action models and eventually data-informed proposal
policies. Controlled comparisons can then measure how each capability affects
success rate, route length, and final physical quality.

\section{Conclusion}

We presented a proof of concept for agentic multi-objective stage-one
stellarator optimization. The language-model agent serves as a stateful
scientific controller over parent equilibria, Fourier-mode access, objective
specifications, and local budgets. A deterministic DESC layer owns the physical
contract, execution, evaluation, and acceptance decision. This division of
responsibility lets the agent conduct extended multi-stage routes and supports
continuous multi-start operation with complete lineage and optimizer feedback.

In a selected common-budget subset of an expanding finite-beta QA campaign, gate-valid
configurations increase from five inputs to nineteen outputs, accompanied by
lower median QS RMS and maximum curvature. A long-route case achieves a
$9.10\times$ QS reduction while repairing magnetic-well and curvature
violations, exposing the nonmonotone acquisition and repair behavior behind the
endpoint. Together, the experiments produce 734 versioned transitions. The
result validates the proposed progression from agentic optimization to data
infrastructure: each route accumulates two coupled assets, physically improved
equilibria and reusable state--action evidence.

The broader proposition is that agentic systems can serve as scientific data
engines for stellarator design. By repeatedly applying expert-like sequential
reasoning through a physics-governed interface, the system can expand the scale,
physical alignment, and informational depth of available design data. This
provides a practical foundation for broader configuration datasets, learned
optimization priors, and increasingly capable agentic design policies.

\FloatBarrier
\scriptsize
\setlength{\bibsep}{0pt}
\bibliographystyle{unsrtnat}
\bibliography{references}

\end{document}